\documentclass[10pt,twocolumn,letterpaper]{article}

\usepackage{cvpr}              
\usepackage{url}
\usepackage{algorithm}
\usepackage{algorithmic}

\usepackage{multirow}
\usepackage[table]{xcolor}
\usepackage{color}
\definecolor{mygreen}{RGB}{146, 199, 113} 
\usepackage{graphicx}
\usepackage{adjustbox}
\usepackage{amsmath}
\usepackage{array}
\usepackage{booktabs}
\usepackage{amssymb}
\usepackage{dsfont}
\usepackage{graphicx}
\usepackage{CJK}

\definecolor{cvprblue}{rgb}{0.21,0.49,0.74}
\usepackage[pagebackref,breaklinks,colorlinks,allcolors=cvprblue]{hyperref}

\def\paperID{xxx} 
\def\confName{CVPR}
\def\confYear{2025}

\title{Redefining \texttt{<Creative>} in Dictionary: \\Towards an Enhanced Semantic Understanding of Creative Generation}

\author{Fu Feng\textsuperscript{\rm 1,2}\quad Yucheng Xie\textsuperscript{\rm 1,2}\quad Xu Yang\textsuperscript{\rm 1,2}\quad Jing Wang\textsuperscript{\rm 1,2}\thanks{Corresponding authors}\quad Xin Geng\textsuperscript{\rm 1,2}\footnotemark[1]\\
\textsuperscript{\rm 1}School of Computer Science and Engineering, Southeast University, Nanjing, China\\
\textsuperscript{\rm 2}Key Laboratory of New Generation Artificial Intelligence Technology and Its Interdisciplinary \\Applications (Southeast University), Ministry of Education, China\\
{\tt\small \{fufeng, xieyc, xuyang\_palm, wangjing91, xgeng\}@seu.edu.cn}\\
}

\begin{document}
\begin{CJK}{UTF8}{gbsn}
\maketitle
\begin{abstract}
``Creative'' remains an inherently abstract concept for both humans and diffusion models. 
While text-to-image (T2I) diffusion models can easily generate out-of-distribution concepts like ``a blue banana'', they struggle with generating combinatorial objects such as ``a creative mixture that resembles a lettuce and a mantis'', due to difficulties in understanding the semantic depth of ``creative''.
Current methods rely heavily on synthesizing reference prompts or images to achieve a creative effect, typically requiring retraining for each unique creative output---a process that is computationally intensive and limits practical applications.
To address this, we introduce CreTok, which brings meta-creativity to diffusion models by redefining ``creative'' as a new token, \texttt{<CreTok>}, thus enhancing models' semantic understanding for combinatorial creativity.
CreTok achieves such redefinition by iteratively sampling diverse text pairs from our proposed CangJie dataset to form adaptive prompts and restrictive prompts, and then optimizing the similarity between their respective text embeddings.
Extensive experiments demonstrate that \texttt{<CreTok>} enables the universal and direct generation of combinatorial creativity across diverse concepts without additional training, achieving state-of-the-art performance with improved text-image alignment and higher human preference ratings.
Code will be made available at~\href{https://github.com/fu-feng/CreTok}{https://github.com/fu-feng/CreTok}.
\end{abstract}

\section{Introduction}
\textit{\small{``Creativity is the power to connect the seemingly unconnected.''}}

\hfill {\textit{\small{--- William Plomer}}}

Recent advancements have witnessed the impressive capabilities of diffusion models, such as DALL-E 3~\cite{ramesh2022hierarchical}, Stable Diffusion 3~\cite{esser2024scaling}, and Midjourney~\cite{midjourney}, which can now generate images comparable to those created by human artists~\cite{balaji2022ediff, peebles2023scalable, zhang2023adding}. 
The high-quality image generation results from models' strength of capturing complex data distributions~\cite{wang2024patch, chen2023score, li2023generalization}. 
However, such strength also leads diffusion models to replicate patterns from their training data, limiting their potential for genuine creativity~\cite{das2022explaining, zhong2024let}.

\begin{figure}[t]
  \centering
  \includegraphics[width=\linewidth]{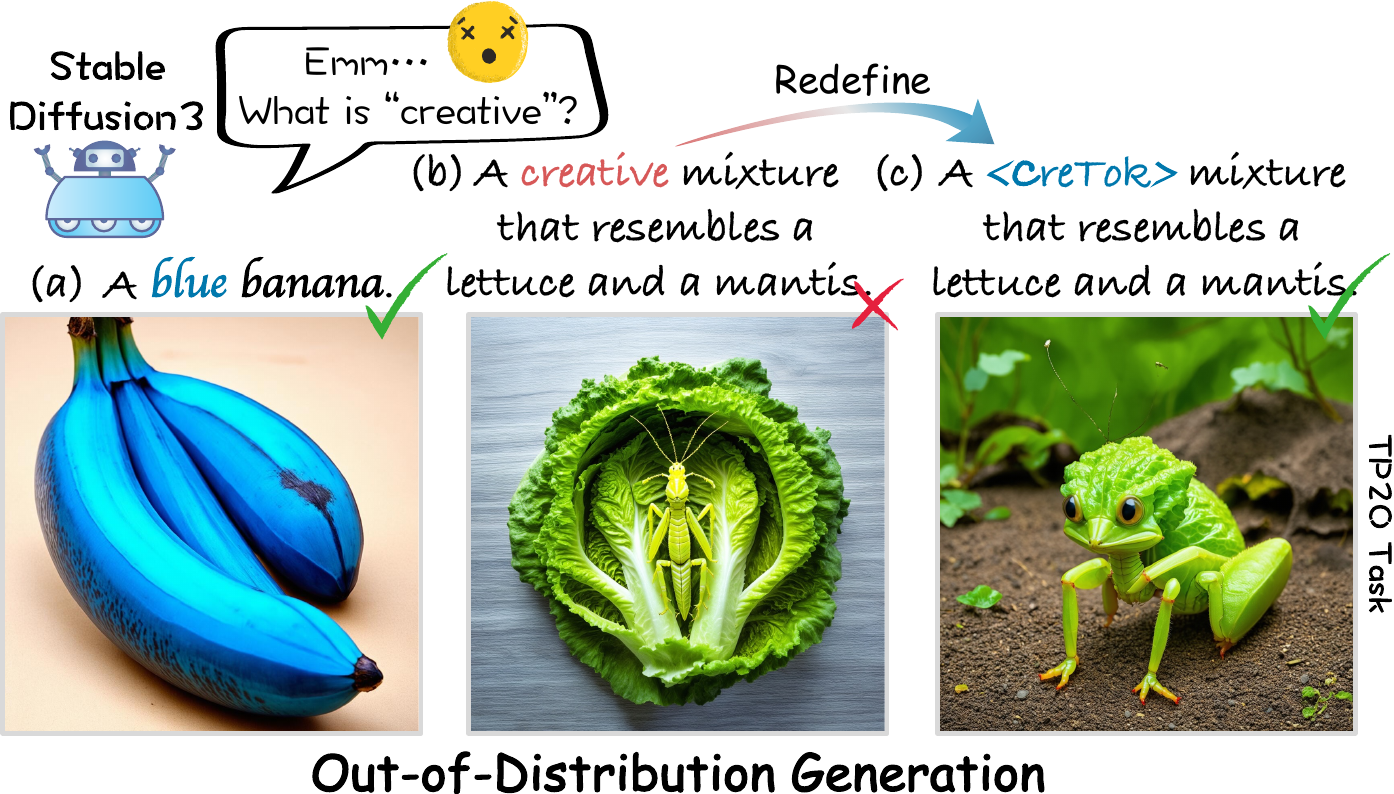}
  \vspace{-0.2in}
  \caption{(a) In out-of-distribution generation, diffusion models can \textbf{\textit{directly}} generate ``a blue banana'' without additional training, benefiting from the clear and concrete semantics of ``blue''.
  (b)\;However, they lack an intrinsic understanding of the abstract and ambiguous semantics of ``creative''.
  (c) Leveraging the TP2O (i.e., Creative Text Pair to Object) task, we redefine the token associated with ``creative'' as \texttt{<CreTok>} to bring models meta-creativity, allowing them to \textbf{\textit{directly}} generate combinatorial creativity by enhancing their semantic understanding of ``creative''.}
  \label{fig:moti}
  \vspace{-0.1in}
\end{figure}

Creativity, an inherently abstract concept even for humans~\cite{sawyer2024explaining, lindstrom2006creativity}, presents a significant challenge for diffusion models. 
Recent efforts have sought to integrate creativity into diffusion models in more concrete ways. For instance, 
ConceptLab~\cite{richardson2024conceptlab} approaches creativity as the generation of novel, indescribable concepts beyond conventional language. 
BASS~\cite{litp2o} extends this interpretation, defining creativity as the \textbf{\textit{combinatorial generation}} of unique objects from text pairs (e.g., \texttt{(Lettuce, Mantis)}), exemplified in Figure~\ref{fig:moti}c. 
Both approaches suggest that creative ability involves generating out-of-distribution images.

Diffusion models can directly generate out-of-distribution images by understanding concrete prompts like ``a blue banana'', but often struggle with the abstract semantics of ``creative'', as illustrated in Figure~\ref{fig:moti}a,b.
Given the challenge that diffusion models face in directly generating creativity, existing methods typically rely on synthesizing reference prompts or images to achieve creative effects.
For instance, to combine ``Lettuce'' and ``Mantis'' creatively, ConceptLab~\cite{richardson2024conceptlab} merges tokens representing these concepts into a new composite token, while BASS~\cite{litp2o} uses predefined sampling rules to search for creative outcomes from a large pool of candidate images.
Similarly, personalization-based methods like MagicMix~\cite{liew2022magicmix}, DiffMorpher~\cite{zhang2024diffmorpher}, and ATIH~\cite{xiong2024novel} use semantic mixing or interpolation during the diffusion process to generate novel visual representations. 

However, these methods rely heavily on reference prompts and images, demanding a new training process for each generation, which leads to high computational costs and limited practicality for online applications. 
In contrast, ``a blue banana'' can be generated directly without additional training, due to its clear and concrete semantics, especially by the adjective ``blue''. 
Inspired by this, we may ask: \textit{Can we awaken the creativity of diffusion models by enhancing their semantic understanding of ``creative''?} 
To achieve this, we propose \textbf{CreTok}, which redefines ``creative'' as a new specialized token, \texttt{<CreTok>}, allowing it to function similarly to ``blue'' in ``a blue banana''. This redefinition enhances the model's semantic understanding for combinatorial creativity, as shown in Figure~\ref{fig:moti}c.
 
Unlike traditional token-based personalization methods, such as textual inversion~\cite{gal2023an, vinker2023concept, ruiz2023dreambooth} and ConceptLab~\cite{richardson2024conceptlab}, which assign a unique token to each static novel concept, CreTok introduces \texttt{<CreTok>} as a \textbf{\textit{universal}} ``adjective'' applicable across all creative concept generation.
Specifically, CreTok builds on the definition of ``creativity'' from the TP2O task~\cite{litp2o} for combinatorial object generation, and refines this concept for meta-creativity in an image-free manner on our proposed dataset of text pairs, termed \textit{CangJie}\footnote{CangJie comes from 仓颉, the creator of Chinese characters.} for the learning of $\texttt{<CreTok>}$.
In each training step, a text pair $(t_1, t_2)$ is randomly sampled to generate creative outputs by optimizing the similarity between the text embedding of a restrictive prompt (e.g., \textit{``A $t_1$ $t_2$''}) and an adaptive prompt (e.g., \textit{``A photo of a $\texttt{<CreTok>}$ mixture}''). 
This process enhances the semantic understanding of $\texttt{<CreTok>}$ for concept-combinatorial creativity beyond the literal meanings of $t_1$ and $t_2$.

Through this approach, CreTok establishes \texttt{<CreTok>} as a universal token that brings \textbf{\textit{meta-creativity}} to diffusion models, transforming creativity from static concept synthesis~\cite{richardson2024conceptlab, litp2o, xiong2024novel} to a more adaptable and flexible creative capability.
This meta-creativity enables the model to generate novel combinatorial concepts, even when the corresponding text pairs have not been encountered during training.
For instance, the combination of \texttt{(Lettuce, Mantis)} in Figure~\ref{fig:moti}c, though unseen during training, can be creatively generated using \texttt{<CreTok>}. 
Furthermore, this meta-creativity enables direct concept combinations without requiring additional training, much like generating ``a blue banana''. 
This significantly reduces both time and computational complexity compared to state-of-the-art (SOTA) creative generation methods, such as ConceptLab~\cite{richardson2024conceptlab} (4s vs. 120s per image, \textbf{30$\times$ speedup}) and BASS~\cite{litp2o} (4s vs. 40s per image, \textbf{10$\times$ speedup}), while maintaining linguistic flexibility for diverse applications and styles. 

Notably, images generated by CreTok achieve higher human preference ratings ($\uparrow$0.009 in PickScore~\cite{kirstain2023pick} and $\uparrow$0.169 in ImageReward~\cite{xu2024imagereward}) and better text-image alignment ($\uparrow$0.03 in VQAScore~\cite{xiong2024novel}) compared to SOTA diffusion models, such as Stable Diffusion 3.5~\cite{stability2024}. 
Further evaluations using GPT-4o~\cite{achiam2023gpt} and user studies  indicate superior performance of CreTok in terms of integration, originality, and aesthetics, underscoring its effectiveness in fostering combinatorial creativity.

Our contributions are as follows:
(1) We propose CreTok, a method designed to enhance models' meta-ability by enabling an enhanced understanding of abstract and ambiguous adjectives (e.g., ``creative'' or ``beautiful'') through their redefinition as new tokens.
(2) Leveraging CreTok, we redefine the abstract term ``creative'' within our proposed \textit{CangJie} dataset for the TP2O task, and introduce \texttt{<CreTok>}, a universal token that imparts meta-creativity to diffusion models, enabling direct application to the creative generation of diverse combinatorial concepts.
(3) Experimental results demonstrate the effectiveness of CreTok in generating combinatorial creativity, outperforming SOTA text-to-image (T2I) models and creative generation methods in terms of computational complexity, human preference ratings, text-image alignment, and other key metrics.

\section{Related Work}
\subsection{Creative Generation}
Advancing machine intelligence necessitates models with human-like creativity, a critical yet underexplored aspect of AI research~\cite{mateja2021towards, mazzone2019art}. 
Early approaches to creativity involves heuristic search methods~\cite{xu2012fit, cohen2016inspired}. With the rise of image generation, interest in exploring creativity expands~\cite{elgammal2017can, heyrani2021creativegan, nobari2022range}, particularly within Generative Adversarial Networks~\cite{goodfellow2020generative} and Variational Autoencoders~\cite{kingma2013auto}.

More recently, text-to-image (T2I) models~\cite{han2024emma, li2025connecting} have incorporated tasks specifically targeting creativity. 
ConceptLab~\cite{richardson2024conceptlab} introduces Creative Text to Image Generation (CT2I) task, which focuses on generating novel visual concepts beyond conventional language description.
In contrast, BASS~\cite{litp2o} proposes the Creative Text Pair to Object (TP2O) task, which combines attributes of existing concepts into new compositions.
Compared to the open-ended nature of CT2I, TP2O offers more controlled, user-aligned creativity.
Parallel advancements in creative text generation~\cite{zhong2024let, singh2023hide} further emphasize the significance of creativity across modalities. 

In this work, we enhance the creativity of diffusion models—particularly in the TP2O task—by refining their semantic understanding of ``creative'' through a redefined token, \texttt{<CreTok>}. 
\texttt{<CreTok>} brings meta-creativity to diffusion models, transforming it into a universal token for expressing ``creative'' and enabling the combination of diverse concepts without additional training.

\subsection{Personalized Visual Content Generation}
Personalization aims to generate diverse images of specific concepts from limited reference images~\cite{ruiz2024hyperdreambooth, peng2024portraitbooth}. 
Foundational approaches, such as Textual Inversion~\cite{gal2023an} and DreamBooth~\cite{ruiz2023dreambooth}, optimize text embeddings to capture unique visual concepts as new tokens. 
Building on these, recent methods employ compositional techniques for creative recombination of visual elements. 
For instance, Concept Decomposition~\cite{vinker2023concept} breaks personalized concepts into distinct visual aspects captured by specific tokens, while PartCraft~\cite{ng2024partcraft} deconstructs images into modular, fine-grained components for selective reassembly.

Beyond text embedding optimization, advanced methods like SVDiff~\cite{han2023svdiff} and others~\cite{zhang2024spectrum, kumari2023multi,chen2024artadapter} enhance model adaptability through targeted network tuning. 
Techniques such as MagicMix~\cite{liew2022magicmix}, DiffMorpher~\cite{zhang2024diffmorpher}, and ATIH~\cite{xiong2024novel} use semantic mixing or interpolation~\cite{zhong2024multi} during the diffusion process to create innovative visual representations.

In this context, \texttt{<CreTok>} functions not as a representation of a specific concept but as a descriptor of \textit{how to generate creativity}, imparting meta-creativity to models and establishing \texttt{<CreTok>} as a universally adaptable token.

\begin{figure*}[t]
  \centering
  \includegraphics[width=\linewidth]{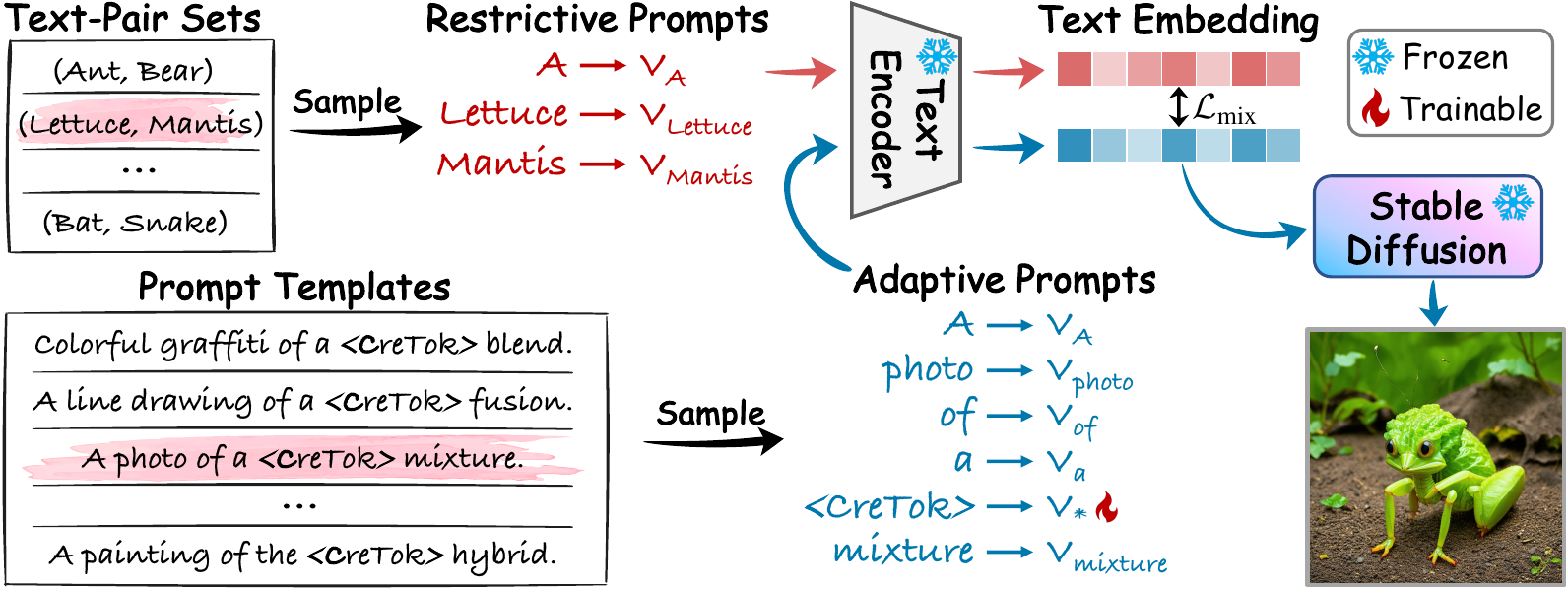}
  \vspace{-0.15in}
  \caption{In each training iteration, a text pair and a prompt template are sampled to create a restrictive prompt and an adaptive prompt. 
  The trainable \texttt{<CreTok>} token is then optimized to minimize the cosine similarity between the text embeddings of the adaptive and restrictive prompt. 
  Then the refined adaptive prompt is input into a diffusion model (e.g., Stable Diffusion 3~\cite{esser2024scaling}) for creative image generation.}
  \label{fig:main}
  \vspace{-0.1in}
\end{figure*}

\section{Methods}
CreTok arouses the creativity of diffusion models by enhancing their semantic understanding of ``creative'' and redefining it as a new token, \texttt{<CreTok>}. 
This section first presents the basic principles of T2I models, followed by a detailed method for combining text pairs into novel concepts. 
Finally, we describe the iterative process to continually refine \texttt{<CreTok>} for enhanced creative expression.

\subsection{Preliminary}
Latent Diffusion Models (LDMs)~\cite{rombach2022high} have been widely adopted in T2I generation~\cite{guo2024smooth, patel2024lambda, mou2024t2i, castells2024ld, yue2024exploring, yue2024few, hu2023cocktail, hu2024one}.
LDMs shift the diffusion process from pixel space to a compact latent space with an encoder $\mathcal{E}$, which maps images $x$ into spatial latent codes $z=\mathcal{E}(x)$.
The diffusion model is then trained to generate these latent codes through denoising, minimizing the following objective:
\begin{equation}
    \mathcal{L}_{\text{diff}}=\mathbb{E}_{z\sim \mathcal{E}(x),y,\varepsilon \sim \mathcal{N}(0,1),t}[||\varepsilon-\varepsilon_{\theta}(z_t, t, c_{\theta}(y))||^{2}_{2}]
\label{eq:loss_LDM}
\end{equation}
where $\varepsilon_{\theta}$ represents the noise prediction network, trained to estimate the noise $\varepsilon$ added to the latent variable $z_t$ at timestep $t$, conditioned on $c_{\theta}(y)$, a vector derived from the input $y$ (e.g., text prompt) through a mapping function $c_{\theta}$.

In our study, we utilize Stable Diffusion 3 (SD 3)~\cite{esser2024scaling} as the base model, which employs a transformer-based architecture to facilitate bidirectional information flow between image and text tokens. 
Our work focuses on the text encoder in $c_{\theta}$ and condition input $y$, while keeping the parameters of other components frozen (See Appendix~\ref{app:SD3} for details).

\subsection{Creative Generation from a Single Text Pair}
To redefine ``creative'' as a universally applicable token, \texttt{<CreTok>}, for the combinatorial generation of various text pairs, we begin by performing token-based concept fusion with a single text pair.
Building on ConceptLab~\cite{richardson2024conceptlab}, we achieve such fusion by increasing the semantic similarity of two distinct prompts in the embedding space.

As shown in Figure~\ref{fig:main}, given a text pair $(t_1, t_2)$ (e.g., \texttt{(Lettuce, Mantis)}), we generate a restrictive prompt $\mathcal{P}_r (t_1, t_2)$ by combining the pair into a phrase like ``a $t_1$ $t_2$'' (e.g., ``a lettuce mantis.''). 
The \textbf{trainable} token $\texttt{<CreTok>}$, which redefines ``creative'', is then used to form an adaptive prompt $\mathcal{P}_a$ representing the combinatorial results (e.g., ``a photo of a $\texttt{<CreTok>}$ mixture.'').
To optimize the semantic alignment of $\mathcal{P}_r$ and $\mathcal{P}_a$, we increase the similarity between the embeddings of $\mathcal{P}_r$ and $\mathcal{P}_a$ using the following objective:
\begin{equation}
    \mathcal{L}_{\text{mix}} = 1 - \text{cos}(E(\mathcal{P}_r (t_1, t_2)), E(\mathcal{P}_a))
\label{eq:loss_gene}
\end{equation}
where $\text{cos}(a, b) = \frac{a\cdot b}{||a||||b||}$ denotes cosine similarity, and $E(\cdot)$ is the text encoder (e.g., CLIP L/14~\cite{radford2021learning} in SD 3) that maps prompts to corresponding text embeddings.

As noted in ConceptLab~\cite{richardson2024conceptlab}, overfitting can artificially inflate similarity by disproportionately reinforcing one concept while neglecting others. To address this issue, we introduce a loss threshold $\theta$ to regulate concept integration.
Moreover, to ensure the coherent fusion of two concepts, rather than their independent generation (e.g., a lettuce and a mantis), $\theta$ must remain moderate to avoid low similarity between $\mathcal{P}_r$ and $\mathcal{P}_a$ (see Section~\ref{sec:threshold} for details).
\begin{equation}
    \tilde{\mathcal{L}}_{\text{mix}} = 1 - \min[\text{cos}(E(\mathcal{P}_r (t_1, t_2)), E(\mathcal{P}_a)), \theta]
\label{eq:threshold}
\end{equation}

Additionally, we observe that the order of $t_1$ and $t_2$ in $\mathcal{P}_r$ can bias the model's subject focus. For instance, ``a lettuce mantis'' may prioritize mantis features with lettuce-like elements. 
To mitigate this bias, we alternate the positions of the two texts in each pair (i.e., $(t_2, t_1)$), and compute the loss for both $\mathcal{P}_r(t_1, t_2)$ and $\mathcal{P}_r(t_2, t_1)$ during training, encouraging a balanced fusion of both concepts. 

\subsection{Refining \textbf{\texttt{<CreTok>}} in a Continuous Process}
Our ultimate objective is not to create a token representing a specific new concept, as done in ConceptLab~\cite{richardson2024conceptlab} and other token-based personalization methods~\cite{gal2023an, ruiz2023dreambooth}. 
Instead, we aim to enhance the diffusion model's semantic understanding of ``creative'' through \texttt{<CreTok>}, thus guiding the model on ``how to generate creativity''.
This meta-creativity cannot be achieved through direct optimization on a single text pair, which risks embedding the specific semantics of $t_1$ and $t_2$ into \texttt{<CreTok>}. 

To achieve such meta-creativity, we construct a dataset specifically for the TP2O task, termed \textit{CangJie}, comprising diverse text pairs (see Appendix~\ref{app:CJ} for details). 
Then we iteratively refine \texttt{<CreTok>} on \textit{CangJie} through a continuous training process, gradually embedding generalized semantics of ``creative'' into \texttt{<CreTok>}.
In each training iteration, a set of $n$ text pairs is randomly sampled, and the cumulative loss is calculated as:
\begin{equation} 
    \mathcal{L}_{\text{iter}} = \frac{\sum_{i=1}^n \tilde{\mathcal{L}}_{\text{mix}}^{i}}{n} 
\end{equation} 
where $\mathcal{L}_{\text{mix}}$ represents the cosine similarity loss between $\mathcal{P}_r$ and $\mathcal{P}_a$, as defined in Eq.~\eqref{eq:loss_gene}.
After each update, new sets of $n$ text pairs are sampled, ensuring that \texttt{<CreTok>} remains generalizable across a wide range of concepts.

\section{Experiments}
\subsection{Datasets}
To comprehensively evaluate creativity, we develop \textit{CangJie}, the first dataset specifically designed for the TP2O task.
\textit{CangJie} combines concepts from categories like animals and plants, forming text pairs through diverse combinations. 
The dataset includes 200 text pairs for training \texttt{<CreTok>}, and 27 text pairs from the original BASS~\cite{litp2o} results for unified comparison. Detailed specifications are provided in Appendix~\ref{app:CJ}.

\subsection{Experimental Setup}
Our implementation is based on the official Stable Diffusion 3~\cite{esser2024scaling}, which integrates three text encoders: CLIP L/14~\cite{radford2021learning}, OpenCLIP bigG/14~\cite{cherti2023reproducible}, and T5-v1.1-XXL~\cite{raffel2020exploring}. 
In our experiments, only the two CLIP models are used as text encoders $E(\cdot)$ without significant performance loss, owing to the simplicity of prompts.
The training of \texttt{<CreTok>} runs for 10K steps on a single NVIDIA 4090 GPU, using an initial learning rate of 0.01 with a cosine scheduler, a batch size of 1, and gradient accumulation over $n=16$ steps.
The training process can be completed within approximately 30 minutes. Notably, there is \textbf{NO} additional computational overhead after the training of \texttt{<CreTok>}. 

\begin{figure*}[t]
  \centering
  \includegraphics[width=\linewidth]{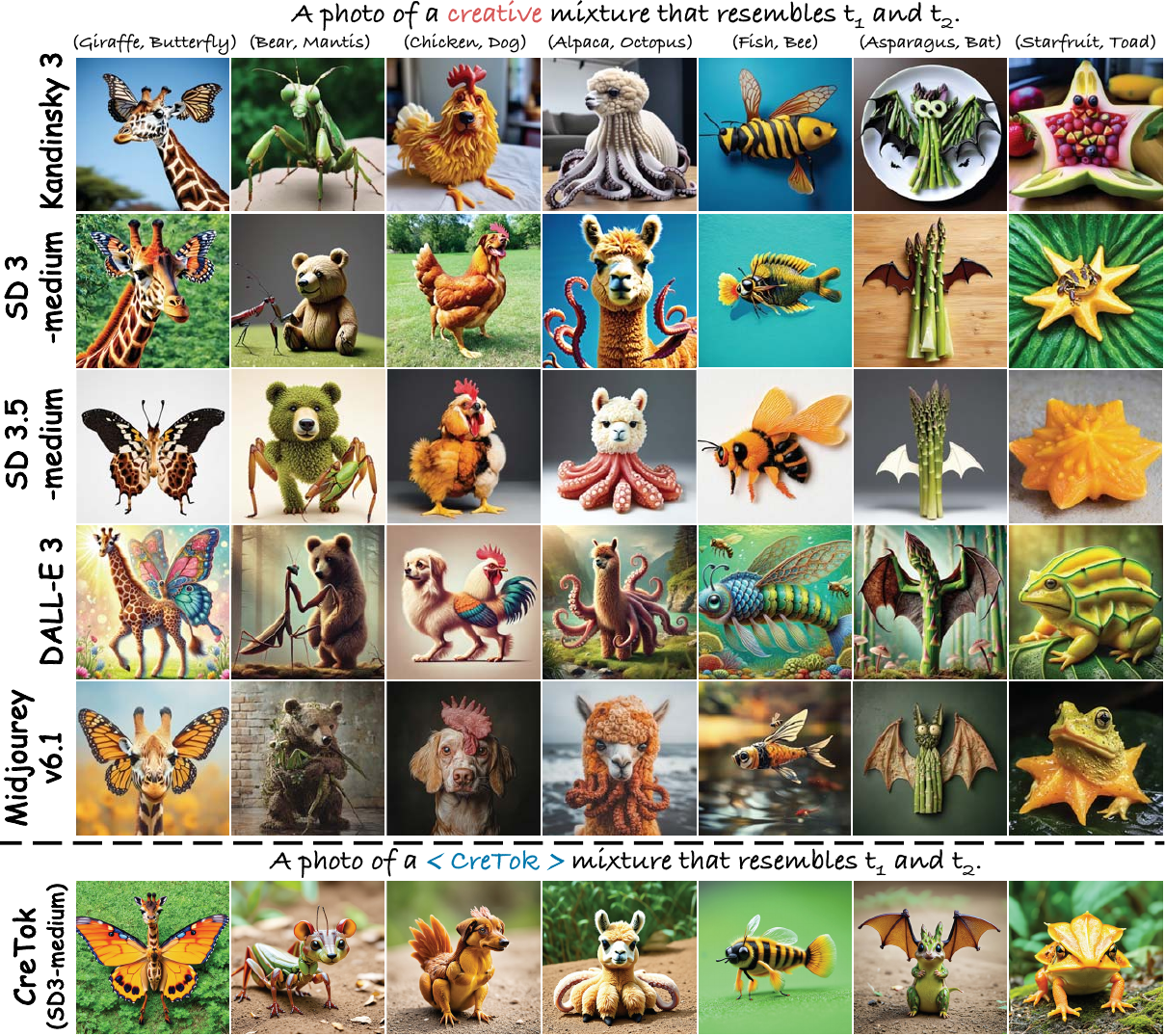}
  \vspace{-0.18in}
  \caption{$\texttt{<CreTok>}$ enhances diffusion models' semantic understanding of combinatorial creativity.
  We compare CreTok with SOTA T2I diffusion models including Stable Diffusion 3~\cite{ramesh2022hierarchical}, Kandinsky 3~\cite{razzhigaev2023kandinsky}, Stable Diffusion 3.5~\cite{stability2024}, DALL-E 3~\cite{ramesh2022hierarchical} and Midjourney v6.1~\cite{midjourney} with identical prompts. CreTok, built on Stable Diffusion 3, replaces ``creative'' in prompts with the redefined \texttt{<CreTok>}.}
  \vspace{-0.1in}
  \label{fig:compare_diff}
\end{figure*}

\subsection{Evaluation Metrics}
To evaluate the creativity generated by CreTok and related methods, we first apply VQAScore~\cite{lin2025evaluating} to measure alignment between the generated image and the text prompt, particularly for combinatorial generation. 
We also employ PickScore~\cite{kirstain2023pick} and ImageReward~\cite{xu2024imagereward} to evaluate alignment with aesthetic standards and human preferences.
Additionally, we use GPT-4o~\cite{openai2023gpt4} and conduct a user study to comprehensively evaluate creativity in terms of conceptual integration, originality, and aesthetic quality.
Generation time per image is recorded to highlight deployment considerations, demonstrating CreTok's zero-shot efficiency.

\begin{figure*}[t]
  \centering
  \includegraphics[width=\linewidth]{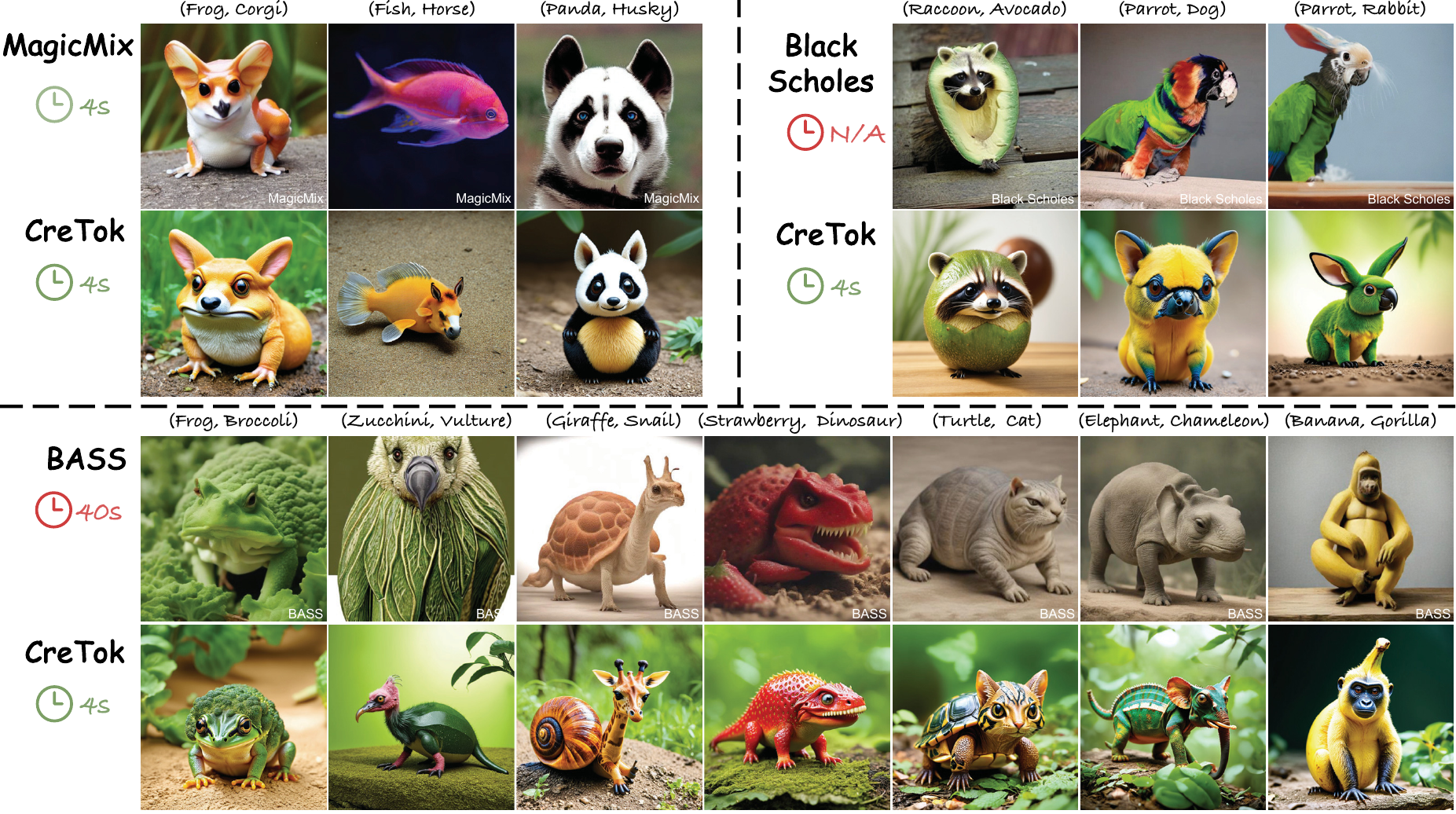}
  \vspace{-0.25in}
  \caption{Visual comparisons of combinatorial creativity. We compare CreTok with BASS~\cite{litp2o}, and other methods achieving similar combinatorial effects, including MagicMix~\cite{liew2022magicmix} and Black-Scholes~\cite{kothandaraman2024prompt}, to highlight CreTok's superior performance. 
  For fair comparison, most images from these methods are sourced directly from the original papers, with a white watermark added in the bottom right corner. 
  Additionally, generation time per image is recorded to emphasize CreTok's meta-creativity and zero-shot capability.}
  \label{fig:zero-shot}
  \vspace{-0.1in}
\end{figure*}

\section{Results}
\subsection{Performance of Redefined \textbf{\texttt{<CreTok>}}}
\subsubsection{Comparison with State-of-the-Art T2I Models}
We evaluate CreTok against state-of-the-art (SOTA) T2I models, including Stable Diffusion 3.5~\cite{stability2024}, DALL-E 3~\cite{ramesh2022hierarchical} and Midjourney v6.1~\cite{midjourney} under identical prompts, as shown in Figure~\ref{fig:compare_diff}. 
Despite extensive training on large-scale datasets, SOTA models still struggle to capture the abstract concept of ``creative'' and struggle to generalize beyond their training distributions, often rendering two objects as separate entities rather than as a cohesive, integrated concept, such as \texttt{(Bear, Mantis)}.

Models like DALL-E 3 and Midjourney demonstrate some improvements in combinatorial generation over Stable Diffusion 3, benefiting from advanced architectures and extensive training. 
However, their outputs often favor an artistic style with vivid colors and intricate details, which contrasts with the realism expected in ``photo'' prompts, making realistic combinatorial generation a more challenging out-of-distribution (OOD) task.

In contrast, CreTok significantly enhances the model's semantic understanding of ``creative'' through \texttt{<CreTok>}, enabling a more cohesive and realistic fusion while preserving the interpretability of each component. Additional images generated by CreTok are available in Appendix~\ref{app:image}.

\subsubsection{Comparison with Creative Generation Methods}
Beyond comparisons with SOTA T2I models, we evaluate CreTok against methods specifically designed for creativity and personalization to further highlight its advantages.
Creative generation methods like BASS~\cite{litp2o} achieve creative outputs through rule-based searches across large-scale candidate generations, while personalization methods, such as MagicMix~\cite{liew2022magicmix} and Black Scholes~\cite{kothandaraman2024prompt}, can generate combination effects via interpolation between noise, prompts.

As shown in Figure~\ref{fig:zero-shot}, interpolation techniques used in personalization can produce similar visual effects, but they are heavily dependent on reference images, limiting their further adaptability. 
When significant visual disparities exist between images, the resulting fusion often lacks coherence (see Appendix~\ref{app:more_compare} for additional comparisons).
While BASS is capable of producing high-quality creative images without reference images, it demands substantial computational resources (40 seconds vs. CreTok's 4 seconds) for generation and filtering.

CreTok addresses limitations by introducing a universal token, \texttt{<CreTok>}, specifically for combinatorial creativity, which directly redefines ``creative'' for meta-creativity rather than merely synthesizing reference images.
This allows \texttt{<CreTok>} to seamlessly integrate with other tokens, facilitating novel concept generation without additional training.

\begin{figure}[t]
  \centering
  \includegraphics[width=\linewidth]{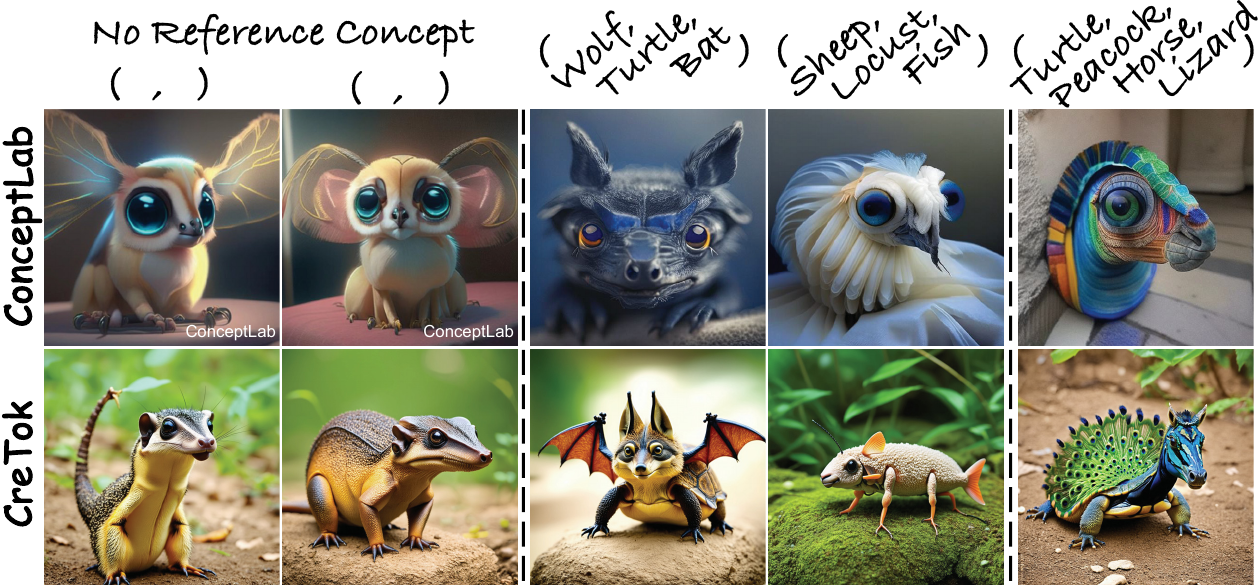}
  \vspace{-0.2in}
  \caption{Combinatorial creativity with no concepts or two more concepts. Images with white watermarks are directly sourced from the original paper of the comparison method.}
  \label{fig:noref}
  \vspace{-0.15in}
\end{figure}

\subsection{Creativity beyond Text Pair}
Beyond text-pair-based TP2O tasks, we extend our exploration to the Creative Text-to-Image (CT2I) task, as defined in ConceptLab~\cite{richardson2024conceptlab}, which allows for the fusion of three or more concepts or the creation of entirely new concepts without referencing existing ones.

Figure~\ref{fig:noref} compares CreTok and ConceptLab on the CT2I task, showcasing creative images generated from multiple or undefined concepts. 
Although \texttt{<CreTok>} is primarily redefined for combinatorial object generation from text pairs, it extends seamlessly to multi-concept fusion, enabling novel creative outputs without reference text. 
For instance, CreTok can generate new concepts using prompts like ``A photo of a $\texttt{<CreTok>}$ mixture.'' without any predefined concepts. 
Moreover, when combining multiple concepts (e.g., \texttt{(Turtle, Peacock, Horse, Lizard)}), ConceptLab struggles to preserve individual concept features, while CreTok effectively maintains the distinct characteristics of each concept.

While ConceptLab also supports multi-concept generation through token updates, each token is tailored to a specific new concept (e.g., ``A photo of $\texttt{<concept>}$''), requiring repeated training for each new creative instance. 
Moreover, CreTok operates directly in CLIP semantic space~\cite{radford2021learning}, without relying on diffusion priors~\cite{ramesh2022hierarchical}, offering a more streamlined framework for creative generation.

\begin{figure}[t]
  \centering
  \includegraphics[width=\linewidth]{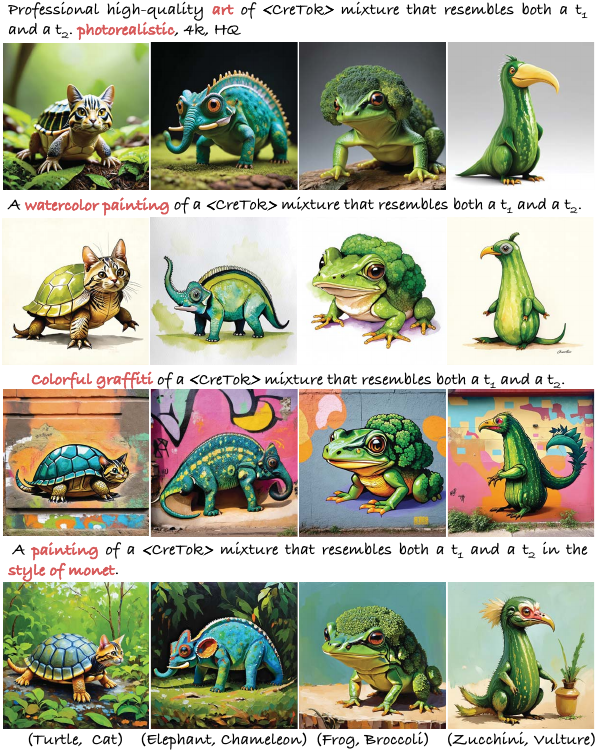}
  \vspace{-0.25in}
  \caption{Redefined \texttt{<CreTok>} can be combined with natural language to showcase combinatorial creativity in various styles.  Additional styles are illustrated in Appendix~\ref{app:style}.}
  \label{fig:prompt}
  \vspace{-0.05in}
\end{figure}

\begin{table}[t]
    \centering
    \setlength{\tabcolsep}{1.3 mm}
    \caption{Quantitative Comparisons for Image-Text Alignment and Human Preference Ratings.}
    \vspace{-0.1in}
    \resizebox{0.47\textwidth}{!}{
        \begin{tabular}{@{}lccc|cc@{}}
        \toprule[1pt]
             & SD 3 & SD 3.5 & Kand 3 & BASS & CreTok \\
             \midrule[0.5pt]
             VQAScore$\uparrow$ & 0.793 & 0.805 & 0.771 & 0.710 & \textbf{0.835}\\
             \midrule[0.5pt]
             PickScore$\uparrow$ & 21.716 & 21.766 & 21.637 & 20.799 & \textbf{21.775} \\
             ImageReward$\uparrow$ & 0.896 & 0.881 & 0.634 & 0.481 & \textbf{1.065} \\
             \bottomrule[1pt]
        \end{tabular}
        }
    \label{tab:human_pre}
    \vspace{-0.07in}
\end{table}

\begin{table}[t]
    \centering
    \setlength{\tabcolsep}{1.5 mm}
    \caption{Creativity evaluated by GPT-4o.}
    \vspace{-0.1in}
    \resizebox{0.47\textwidth}{!}{
        \begin{tabular}{@{}lccccc@{}}
        \toprule[1pt]
             & Integ. & Align. & Orig. & Aesth. & \textit{Compr.} \\
             \midrule[0.5pt]
             SD 3~\cite{esser2024scaling} & 8.1$\pm$4.1 & 8.7$\pm$4.0 & 8.2$\pm$4.1 & 9.0$\pm$1.3 & \textit{8.5$\pm$3.1} \\
             Kand 3~\cite{razzhigaev2023kandinsky} & 8.9$\pm$0.8 & 9.7$\pm$0.3 & 9.0$\pm$0.4 & 9.2$\pm$0.2 & \textit{9.2$\pm$0.3} \\
             SD 3.5 & 9.1$\pm$0.7 & 9.9$\pm$0.2 & 9.1$\pm$0.6 & 9.4$\pm$0.4 & \textit{9.4$\pm$0.3} \\
             \midrule
             BASS~\cite{litp2o} & 8.9$\pm$1.3 & 9.3$\pm$1.4 & 8.7$\pm$1.2 & 8.3$\pm$0.7 & \textit{8.8$\pm$0.9} \\
             CreTok & \textbf{9.5$\pm$0.4} & \textbf{9.9$\pm$0.1} & \textbf{9.3$\pm$0.4} & \textbf{9.6$\pm$0.3} & \textbf{\textit{9.6$\pm$0.3}} \\
             \bottomrule[1pt]
        \end{tabular}
    }
    \label{tab:gpt}
    \vspace{-0.1in}
\end{table}

\subsection{Universality of \textbf{\texttt{<CreTok>}} Among Styles}
A key limitation of existing methods is their inability to transfer generated creativity across various styles.
As previously discussed, \texttt{<CreTok>} serves as a universal ``adjective'',  functioning similarly to ``blue'', allowing it to be seamlessly combined with other prompts for various styles, such as ``painting'' or ``art''. 
Figure~\ref{fig:prompt} presents our results across diverse image styles, highlighting CreTok's unique adaptability—a capability that cannot be achieved by methods like MagicMix~\cite{liew2022magicmix} and BASS~\cite{litp2o}. 

Unlike ConceptLab~\cite{richardson2024conceptlab}, where each token is tied to a specific concept, \texttt{<CreTok>} does not correspond directly to any single concept, yet it consistently maintains adaptability across a wide range of prompts.

\begin{figure*}[t]
  \centering
  \includegraphics[width=\linewidth]{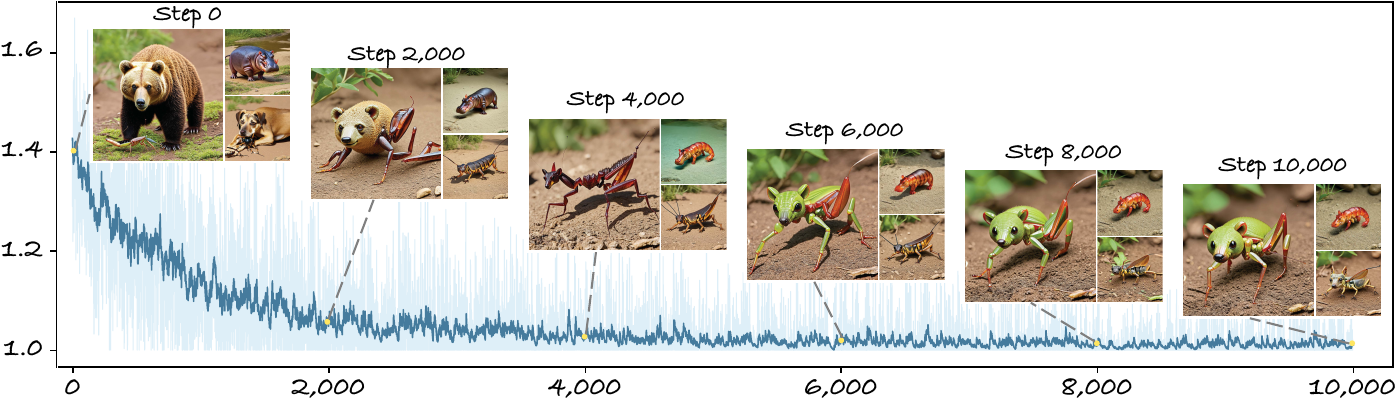}
  \vspace{-0.28in}
  \caption{Convergence rate of \texttt{<CreTok>} during the continuous redefinition process, showing training curves and corresponding images.}
  \label{fig:loss}
  \vspace{-0.15in}
\end{figure*}

\subsection{Evaluation for Creativity}
\subsubsection{Quantitative Comparisons}
We conduct quantitative comparisons to evaluate the alignment between images and prompts using VQAScore~\cite{lin2025evaluating}, along with human preference ratings via PickScore~\cite{kirstain2023pick} and ImageReward~\cite{xu2024imagereward}. 
Table~\ref{tab:human_pre} presents comparisons between CreTok and SOTA open-source T2I models.

Despite being built upon SD 3, CreTok outperforms both SD 3.5 and Kandinsky 3 in terms of human preference ratings and image-text alignment, even though these models use advanced architectures and extensive training data tailored to human aesthetic preferences.

\subsubsection{Evaluation via GPT-4o}
Since existing metrics are insufficient for assessing such abstract ``creativity'', we employ GPT-4o to objectively assess image creativity through quantitative analysis across four dimensions: Integration, Alignment, Originality, and Aesthetics. 
Detailed prompts are available in Appendix~\ref{app:prompt}.

Table~\ref{tab:gpt} presents GPT-4o's assessments of creativity for images generated by CreTok compared to other methods. 
The results indicate that CreTok-generated images demonstrate significant advantages across all evaluated dimensions, especially in the concept integration and originality.

\subsubsection{User Study}
To comprehensively evaluate creativity, we conduct a user study involving 50 highly educated participants. Each participant ranks the creativity of images generated by CreTok in comparison with other methods.
The average ranks, summarized in Table~\ref{tab:user_study}, reveal that CreTok significantly outperforms the current T2I diffusion models lacking specialized design for creativity and receives a higher ranking than BASS~\cite{litp2o}, achieving an average ranking of 1.9. Further details are provided in Appendix~\ref{app:user}.

\begin{table}
    \centering
    \setlength{\tabcolsep}{0.8 mm}
    \caption{Results of the user study.}
    \vspace{-0.1in}
    \resizebox{0.47\textwidth}{!}{
        \begin{tabular}{@{}lccccc@{}}
        \toprule[1pt]
             & SD 3 & SD 3.5 & Kand 3 & BASS & CreTok \\
             \midrule[0.5pt]
             Avg. Rank$\downarrow$ & 3.4$\pm$1.5 & 3.1$\pm$1.1 & 3.3$\pm$1.4 & 3.1$\pm$1.3 & \textbf{1.9}$\pm$\textbf{1.1}\\
             \bottomrule[1pt]
        \end{tabular}
        }
    \label{tab:user_study}
    \vspace{-0.15in}
\end{table}

\section{Ablation and Analysis}
\subsection{Process of Continual Redefinition}
The continual refinement of ``creative'' within CreTok is illustrated in Figure~\ref{fig:loss}, which captures the convergence process of $\texttt{<CreTok>}$ over time. 
Additionally, visualizations of randomly selected text pairs, captured every 2,000 training steps, demonstrate the evolving representation.

In early stages, \texttt{<CreTok>} primarily absorbs semantic content from individual concepts, as observed at steps 2,000 and 4,000, where the generated creative output closely resembles one of the concepts (e.g., ``Bear'' and ``Mantis''). 
However, as training progresses, \texttt{<CreTok>} transitions toward encapsulating a generalized creative representation, independent of specific concepts. 
This transition is evidenced by increasingly aligned text-image relationships and enhanced image quality, culminating in final convergence.

\begin{figure}[t]
  \centering
  \includegraphics[width=\linewidth]{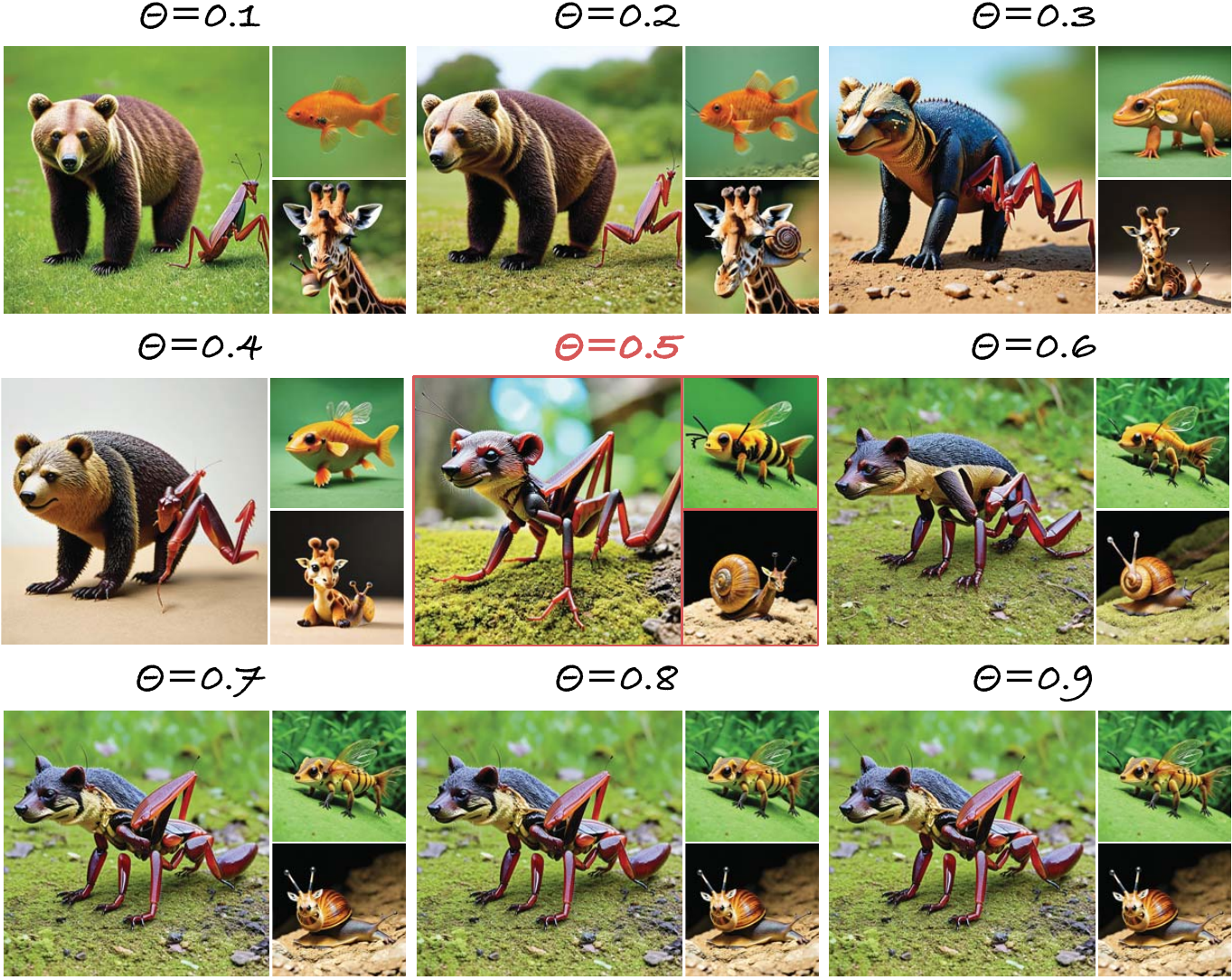}
  \vspace{-0.25in}
  \caption{Combinatorial creativity with different threshold $\theta$.}
  \label{fig:threshold}
  \vspace{-0.18in}
\end{figure}

\subsection{Effect of Loss Threshold on Creativity}
\label{sec:threshold}
When optimizing semantic similarity between text embeddings of restrictive and adaptive prompts, improper thresholds can hinder the combinatorial generation of two concepts. 
To evaluate this, we analyze different thresholds during the refinement of \texttt{<CreTok>}, as shown in Figure~\ref{fig:threshold}. 

Low semantic similarity results in the two concepts being generated independently, rather than merging into a cohesive combination, as demonstrated by current T2I model in Figure~\ref{fig:compare_diff}. 
On the other hand, high similarity increases the likelihood of overfitting to one of the concepts.
Therefore, we identify an optimal threshold of $\theta=0.5$, which strikes a balance between capturing semantic representations and promoting combinatorial object generalization.

\section{Conclusion}
We propose CreTok, a novel approach that imparts meta-creativity to T2I diffusion models by enhancing their semantic understanding of ``creative''. 
CreTok achieves this by redefining ``creative'' as a universal token, \texttt{<CreTok>}, enabling the model to achieve combinatorial creativity in a zero-shot, image-free manner. 
Moreover, \texttt{<CreTok>} integrates seamlessly with natural language, facilitating concept combinations across various styles without additional training.
Extensive experiments demonstrate that CreTok significantly enhances model creativity, outperforming SOTA T2I models and creative generation methods.

\section*{Acknowledgement}
We sincerely thank Wenqian Li, Jianlu Shen, and Ruixiao Shi for their insightful discussions on this work. We also appreciate Freepik for contributing to the figure design. 
This research is supported by the Science and Technology Major Project of Jiangsu Province under Grant BG20240305, Key Program of Jiangsu Science Foundation under Grant BK20243012, National Natural Science Foundation of China under Grants U24A20324, 62125602, and 62306073, Natural Science Foundation of Jiangsu Province under Grant BK20230832, and the Xplorer Prize.

{
    \small
    \bibliographystyle{ieeenat_fullname}
    \bibliography{main}
}

\clearpage
\appendix
\counterwithin{figure}{section}
\counterwithin{table}{section}
\counterwithin{equation}{section}

\section{More Details on Stable Diffusion 3}
\label{app:SD3}
The text encoder in Stable Diffusion 3 incorporates three language models: CLIP L/14 model $c_{\theta}^{\text{\tiny{CLIP-L}}}(\cdot)$\cite{radford2021learning}, OpenCLIP bigG/14 model $c_{\theta}^{\text{\tiny{CLIP-G}}}(\cdot)$\cite{cherti2023reproducible}, and T5-v1.1-XXL model $c_{\theta}^{\text{\tiny{T5}}}(\cdot)$~\cite{raffel2020exploring}. 
Due to the simplicity of prompts in our work, the T5 model is omitted without performance loss.
For a prompt $y$, the CLIP-based encoders $c_{\theta}^{\text{\tiny{CLIP-L}}}(\cdot)$ and $c_{\theta}^{\text{\tiny{CLIP-G}}}(\cdot)$ generate corresponding text embeddings, $c_{\theta}^{\text{\tiny{CLIP-L}}}(y)$ and $c_{\theta}^{\text{\tiny{CLIP-G}}}(y)$.

These embeddings are concatenated post-pooling to form a vector conditioning $c_{\text{vec}}\in \mathbb{R}^{2048}$. Additionally, penultimate hidden layer representations from each model are concatenated along the channel dimension, producing a CLIP context conditioning $c_{\text{ctxt}}\in \mathbb{R}^{77\times 2048}$. Our method then exclusively applies operations to $c_{\text{vec}}$.

\section{Additional Results}
\begin{figure*}[!h]
  \centering
  \includegraphics[width=\linewidth]{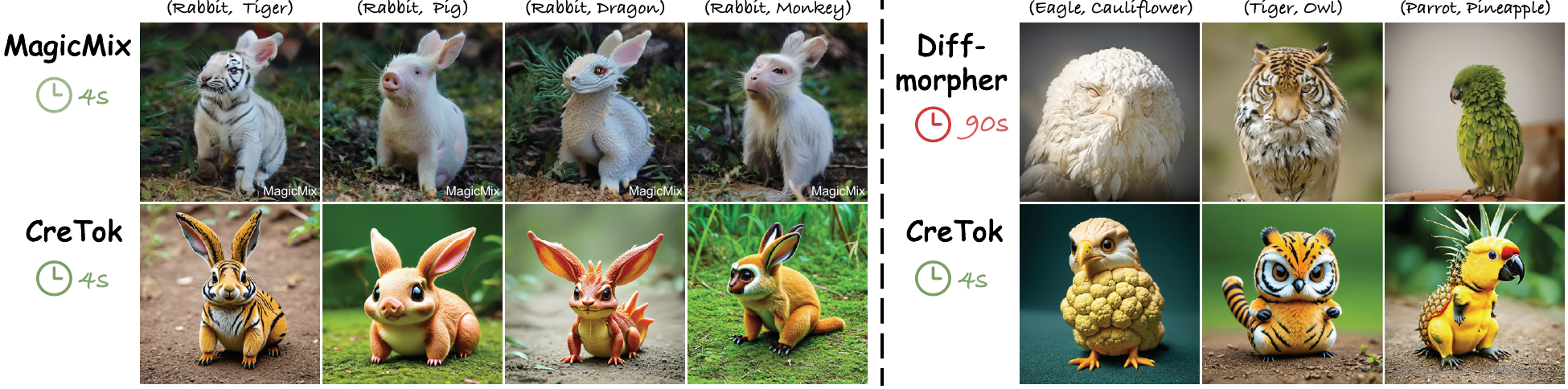}
  \vspace{-0.2in}
  \caption{More visual comparisons of combinatorial creativity. We further compare CreTok with MagicMix~\cite{liew2022magicmix} and DiffMorpher~\cite{zhang2024diffmorpher}, to further highlight CreTok's superior performance. }
  \label{fig:app_compare}
\end{figure*}

\subsection{Additional Comparison Results}
\label{app:more_compare}
We further compare CreTok with other personalization methods designed to produce combination effects through interpolation, such as MagicMix~\cite{liew2022magicmix} and Diffmorpher~\cite{zhang2024diffmorpher}.

As illustrated in Figure~\ref{fig:app_compare}, interpolation-based techniques heavily rely on reference images, which constrains their adaptability.
When substantial visual differences exist between the input images, the resulting fusion often appears incoherent, a limitation particularly evident in Diffmorpher.

In contrast, CreTok enables the diffusion model to directly generate combinatorial creativity without relying on image synthesis, producing outputs with enhanced visual coherence and detail.

\subsection{Additional Creative Visual Results}
\label{app:image}
\begin{figure*}[t]
  \centering
  \includegraphics[width=\linewidth]{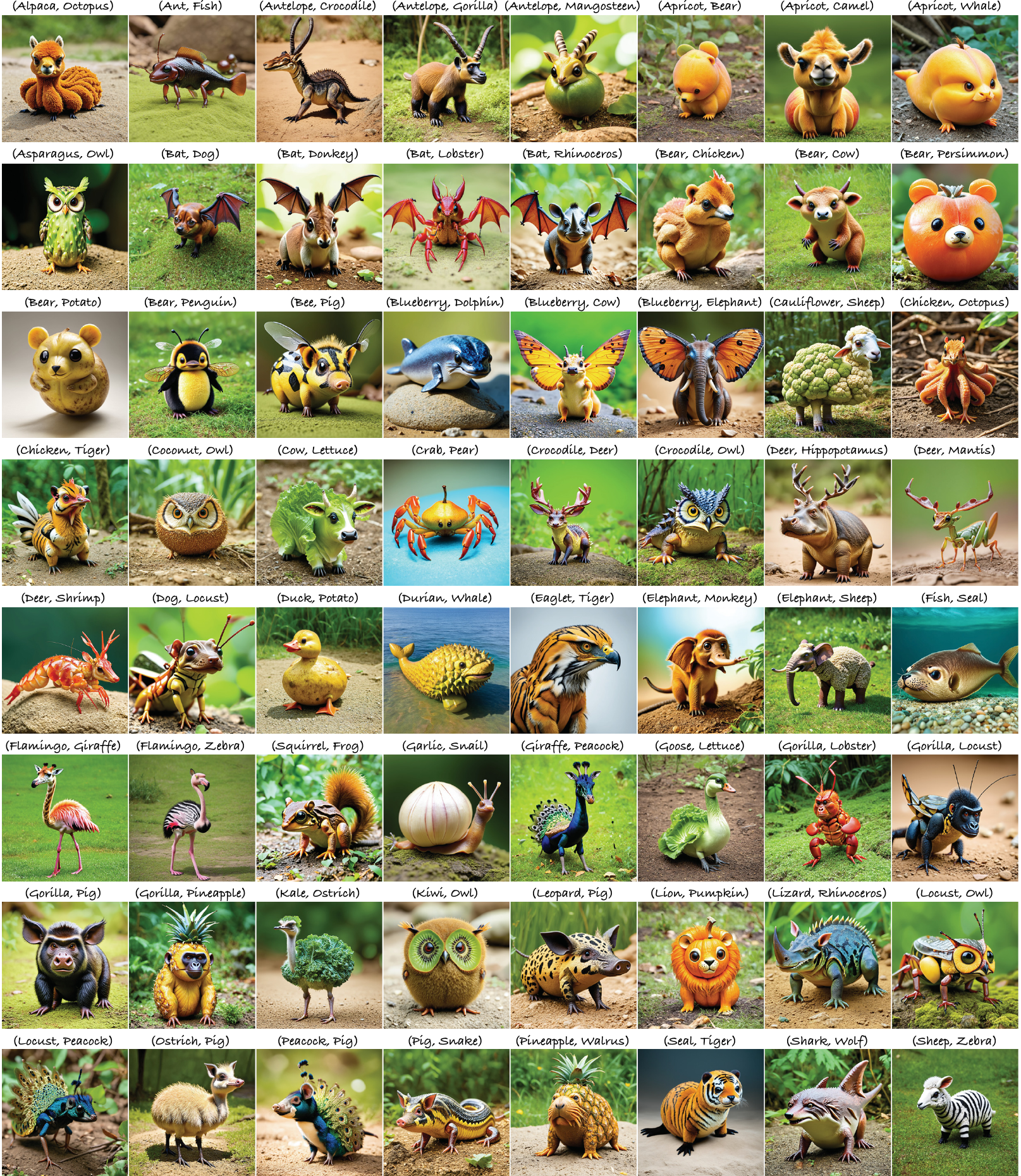}
  \caption{Additional results of combinatorial creativity generated by CreTok.}
  \label{fig:app_more}
\end{figure*}

We present additional creative images generated by CreTok in Figure~\ref{fig:app_more}, further demonstrating the universality of \texttt{<CreTok>} in imparting meta-creativity to diffusion models. 
For each text pair ($t_1$, $t_2$), the corresponding combinatorial creativity is produced using the prompts: ``A photo of a \texttt{<CreTok>} mixture that resembles $t_1$ and $t_2$''. 

\subsection{Additional Styles of Creativity}
\label{app:style}
\begin{figure*}[t]
  \centering
  \vspace{-0.2in}
  \includegraphics[width=\linewidth]{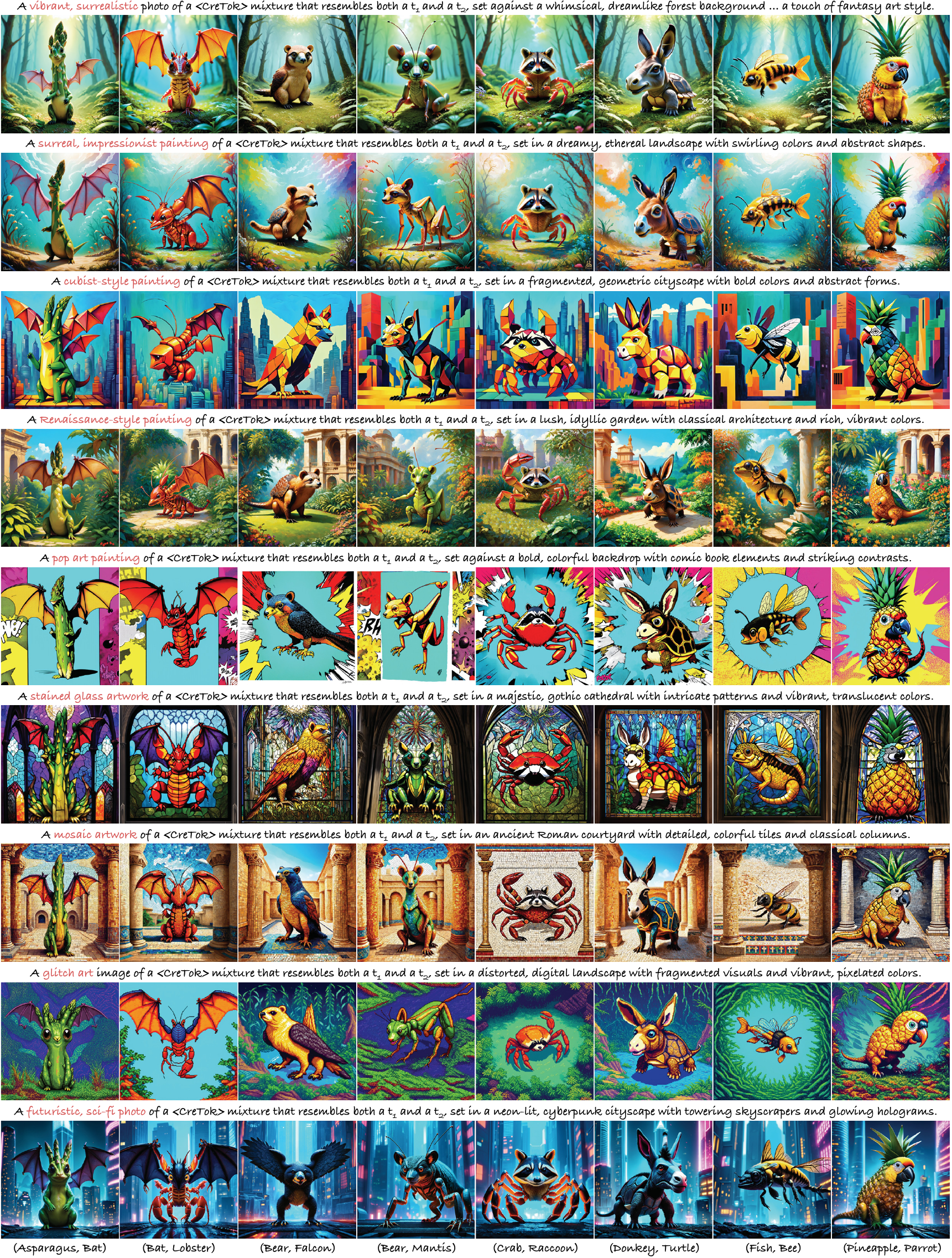}
  \caption{More styles of combinatorial creativity generated by CreTok.}
  \label{fig:app_style}
\end{figure*}

In addition to the creative image styles presented in Figure~\ref{fig:prompt}, Figure~\ref{fig:app_style} showcases additional results, further illustrating the universality of \texttt{<CreTok>}. These results further highlight its seamless integration with natural language, enabling diverse and flexible concept combinations across various styles.

\section{More Details on Evaluation}
\subsection{Prompts Used for GPT-4o Evaluation}
\label{app:prompt}
We conduct an objective evaluation of the creativity of images generated by CreTok and other methods using GPT-4o, assessing four key dimensions: Conceptual Integration, Alignment with Prompt, Originality, and Aesthetic Quality. 
The detailed prompts provided to GPT-4o are as follows:

\textit{The subject of this evaluation is an image that represents a mixture of a banana and a gorilla. The objective is to assess the creativity of an entity that synthesizes two distinct concepts as delineated in the provided prompt. Accordingly, please evaluate the creativity of images generated by various methodologies for the identical prompt, utilizing the following criteria on a scale from 1 to 10:}

\textit{1. Conceptual Integration (1-10): This criterion gauges the degree to which the image manifests a coherent and integrated concept, as opposed to merely placing two independent elements side by side. A high score signifies that the elements are intricately merged, creating a new, unified entity.}

\textit{2. Alignment with Prompt (1-10): This evaluates the extent to which the image conforms to and encapsulates the specific combination of concepts described in the prompt. The image should refrain from including irrelevant elements that detract from the primary concepts. A high score is allocated when the image closely adheres to the specifications of the prompt.}

\textit{3. Originality (1-10): This assesses the innovativeness of the concept portrayed in the image. The depicted concept should not mimic existing animals, plants, or widely recognized mythical creatures unless specifically mentioned in the prompt. Images that present a distinctive and inventive amalgamation receive a high score.}

\textit{4. Aesthetic Quality (1-10): This criterion scrutinizes the visual appeal of the image, focusing on color harmony, the balance and arrangement of elements, and the overall visual impact. A high score is awarded when the image is not only conceptually robust but also visually engaging.}

\textit{In conclusion, based on the aforementioned criteria, provide a comprehensive creative assessment (1-10) and articulate specific justifications for your rating.}

\subsection{More Details on User Study}
\label{app:user}
\begin{figure}[t]
  \centering
  \includegraphics[width=\linewidth]{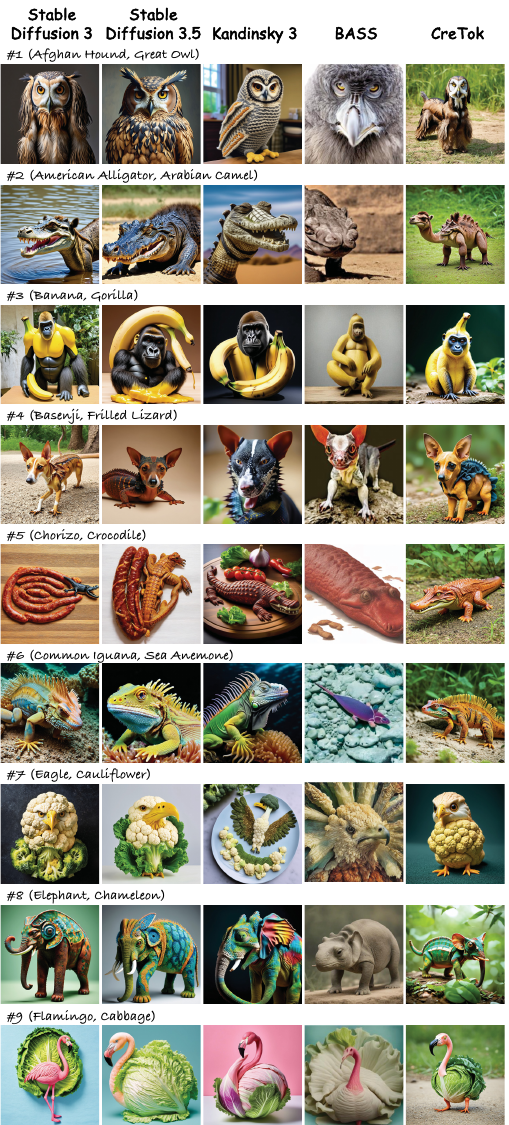}
  \caption{Images generated by CreTok and other methods used in the User Study.}
  \label{fig:app_user1}
\end{figure}

\begin{figure}[t]
  \centering
  \includegraphics[width=\linewidth]{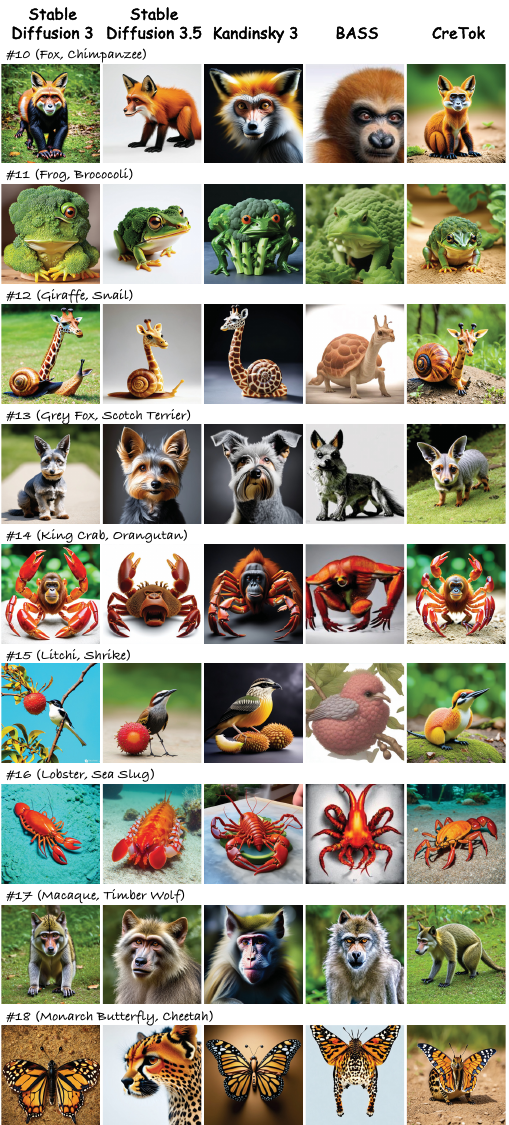}
  \caption{Images generated by CreTok and other methods used in the User Study (continued).}
  \label{fig:app_user2}
\end{figure}

\begin{figure}[t]
  \centering
  \includegraphics[width=\linewidth]{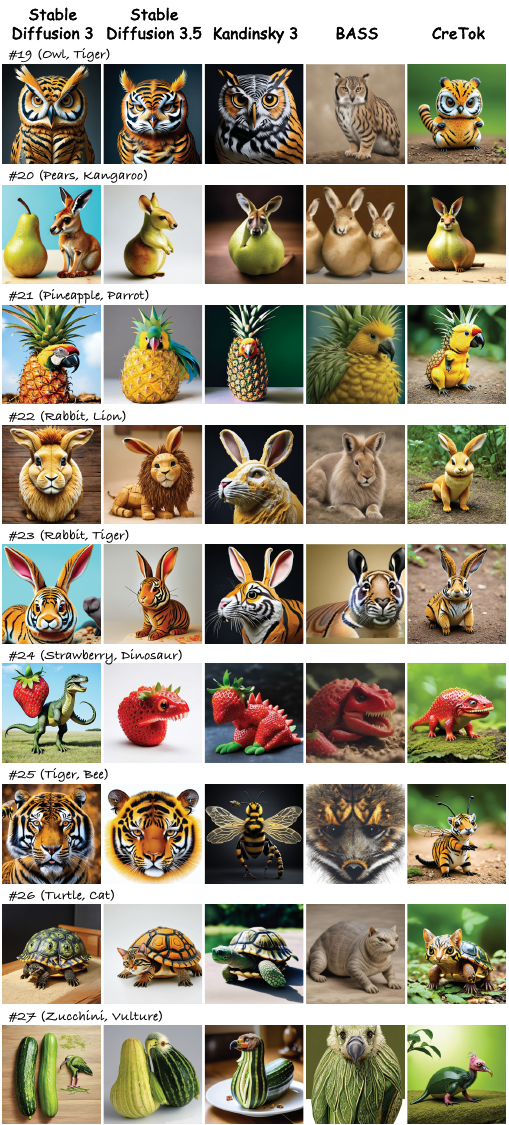}
  \caption{Images generated by CreTok and other methods used in the User Study (continued).}
  \label{fig:app_user3}
\end{figure}

\begin{figure}[t]
  \centering
  \includegraphics[width=\linewidth]{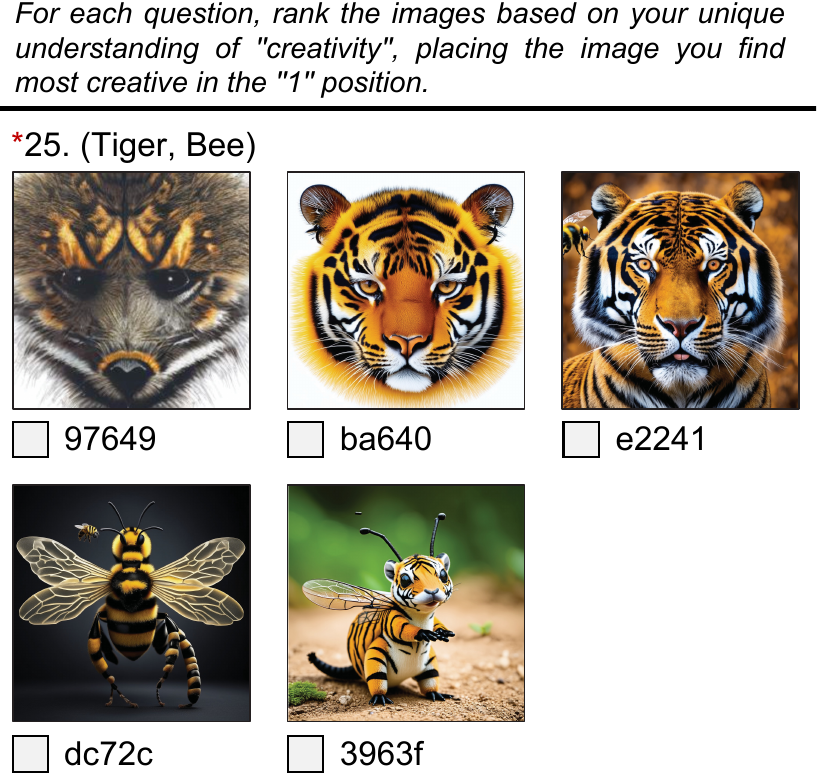}
  \caption{Interface of the User Study.}
  \label{fig:app_infer}
\end{figure}

This section provides a detailed overview of our User Study. The interface used for the study is illustrated in Figure~\ref{fig:app_infer}. We utilize 27 text pairs sourced from the original BASS paper to ensure a fair comparison. 
Creative images generated by CreTok and other methods are displayed in Figure~\ref{fig:app_user1}, \ref{fig:app_user2}, and \ref{fig:app_user3}. This user study involves 50 participants who evaluate and rank the creative outputs for each text pair (1-5). Their responses are summarized in Table~\ref{tab:app_user}.

\begin{table*}[t]
    \centering
    \setlength{\tabcolsep}{0.8 mm}
    \caption{Average ranking of each text pair evaluated in the User Study.}
    \resizebox{\textwidth}{!}{
        \begin{tabular}{@{}lllllllllllllll@{}}
        \toprule[1pt]
             & \texttt{\#1} & \texttt{\#2} & \texttt{\#3} & \texttt{\#4} & \texttt{\#5} & \texttt{\#6} & \texttt{\#7} & \texttt{\#8} & \texttt{\#9} & \texttt{\#10} & \texttt{\#11} & \texttt{\#12} & \texttt{\#13} & \texttt{\#14}\\
        \midrule[0.5pt]
             Stable Diffusion 3 & 1.86$^*$ & 2.50$^*$ & 2.74$^*$ & 4.04 & 3.74 & \textbf{2.26} & \textbf{2.22} & 3.04 & 4.58 & 2.62$^*$ & 2.98$^*$ & 3.66 & \textbf{2.54} & 2.84\\
             Stable Diffusion 3.5 & 3.52 & 4.20 & 3.76 & \textbf{1.98} & 3.02 & 3.02 & 2.48$^*$ & 2.96 & 2.88 & 3.50 & 3.00 & 2.88 & 2.90 & 3.60\\
             Kandinsky 3 & 3.90 & 2.64 & 3.40 & 2.70$^*$ & 2.68$^*$ & 3.00 & 3.78 & \textbf{2.26} & \textbf{1.70} & 2.86 & 3.30 & \textbf{2.14} & 3.82 & \textbf{2.54}\\
             BASS & 3.90 & 3.82 & 2.98 & 3.52 & \textbf{2.08} & 3.76 & 2.86 & 4.38 & 3.44 & 3.70 & 2.98 & 3.64 & 2.92 & 3.44 \\
             CreTok & \textbf{1.82} & \textbf{1.84} & \textbf{2.12} & 2.76 & 3.48 & 2.96$^*$ & 3.66 & 2.36$^*$ & 2.40$^*$ & \textbf{2.32} & \textbf{2.74} & 2.68$^*$ & 2.82$^*$ & 2.58$^*$\\
        \bottomrule[1pt]
        \toprule[1pt]
            & \texttt{\#15} & \texttt{\#16} & \texttt{\#17} & \texttt{\#18} & \texttt{\#19} & \texttt{\#20} & \texttt{\#21} & \texttt{\#22} & \texttt{\#23} & \texttt{\#24} & \texttt{\#25} & \texttt{\#26} & \texttt{\#27} & \\
        \midrule[0.5pt]
            Stable Diffusion 3 & 4.28 & 3.16 & 3.00 & 3.78 & 2.92 & 4.20 & 3.40 & 2.54$^*$ & 2.56$^*$ & 4.64 & 4.18 & 3.78 & 4.36 &\\
            Stable Diffusion 3.5 & 3.90 & \textbf{1.88} & 3.00 & 3.32 & 2.56$^*$ & 2.12$^*$ & 3.22 & \textbf{2.30} & 2.76 & 3.04 & 3.62 & 1.96$^*$ & 3.04 &\\
            Kandinsky 3 & 3.08 & 3.70 & 3.58 & 3.50 & 3.98 & 3.32 & 4.14 & 3.66 & 3.40 & 2.52$^*$ & 2.96 & 4.28 & 2.68$^*$ &\\ 
            BASS & \textbf{1.56} & 3.14 & \textbf{2.64} & \textbf{2.14} & 3.44 & 3.38 & 2.20$^*$ & 2.74 & 3.74 & 2.62 & 2.94$^*$ & 3.28 & \textbf{2.20} &\\
            CreTok & 2.18$^*$ & 3.12$^*$ & 2.78$^*$ & 2.26$^*$ & \textbf{2.10} & \textbf{1.98} & \textbf{2.04} & 3.76 & \textbf{2.54} & \textbf{2.18} & \textbf{1.30} & \textbf{1.70} & 2.72 &\\
        \bottomrule[1pt]
        \end{tabular}
        }
    \label{tab:app_user}
\end{table*}

\section{More Details on Proposed \textit{CangJie} Dataset}
\label{app:CJ}
We introduce \textit{CangJie}, the first dataset specifically designed for the combinatorial creativity proposed in the TP2O task. 
Named after 仓颉, the creator of Chinese characters, \textit{CangJie} symbolizes the dataset's focus on generating novel concepts.
\textit{CangJie} comprises 200 text pairs, combining common animals (e.g., dogs, cats, pigs) and plants (e.g., bananas, pineapples, lettuce). Detailed composition of the dataset are presented in Table~\ref{tab:app_dataset}.
\renewcommand{\arraystretch}{1}
\begin{table*}[t]
    \centering
    \setlength{\tabcolsep}{1.7 mm}
    \caption{Deatils of \textit{CangJie}.}
    \vspace{-0.1in}
    \resizebox{\textwidth}{!}{
        \begin{tabular}{@{}lllll@{}}
        \toprule[1pt]
             (Alpaca, Lion) & (Alpaca, Zebra) & (Ant, Cat) & (Ant, Deer) & (Ant, Gibbon)\\
             (Ant, Horse) & (Ant, Rhinoceros) & (Antelope, Gibbon) & (Antelope, Gorilla) & (Antelope, Rhinoceros) \\
             (Antelope, Walrus) & (Antelope, Zebra) & (Apricot, Bear) & (Asparagus, Owl) & (Banana, Giraffe) \\
             (Banana, Seal) & (Bat, Dog) & (Bat, Donkey) & (Bat, Grape) & (Bat, Lobster)\\
             (Bat, Mantis) & (Bat, Octopus) & (Bat, Rhinoceros) &  (Bat, Zebra) & (Bear, Elephant)\\
             (Bear, Gorilla) & (Bear, Persimmon) & (Bear, Zebra) & (Bee, Carrot) & (Bee, Fish) \\
             (Bee, Frog) & (Bee, Lizard) & (Bee, Penguin)  & (Bee, Pig) & (Bee, Tiger) \\
             (Beetroot, Crab) & (Beetroot, Eagle) & (Beetroot, Octopus) & (Blackberry, Spider) & (Blueberry, Seal) \\(Butterfly, Chicken) & (Butterfly, Elephant) & (Butterfly, Fox) & (Butterfly, Giraffe) & (Butterfly, Parrot)\\(Butterfly, Squid) & (Butterfly, Zucchini) & (Carrot, Lizard) & (Cat, Elephant) & (Cat, Squirrel)\\
             (Cat, Toad) & (Cat, Zebra) & (Cauliflower, Sheep) & (Cherry, Owl) & (Chicken, Leopard)\\
             (Chicken, Octopus) & (Chicken, Pineapple) &(Chicken, Tiger) & (Coconut, Owl) & (Cow, Giraffe)\\
             (Cow, Hippopotamus) & (Cow, Lion) & (Cow, Lobster) & (Cow, Zebra) & (Crab, Pear) \\
             (Crab, Potato) & (Crab, Raccoon) & (Crab, Sheep) & (Cricket, Cucumber) & (Cricket, Fish)\\
             (Cricket, Leopard) & (Cricket, Lizard) & (Cricket, Octopus) & (Cricket, Shark) & (Cricket, Squid) \\
             (Crocodile, Rhinoceros) & (Crocodile, Turtle) & (Crocodile, Zebra) & (Deer, Dog) & (Deer, Fox) \\
             (Deer, Gibbon) & (Deer, Hippopotamus) & (Deer, Monkey) & (Deer, Whale) & (Deer, Wolf) \\
             (Deer, Zucchini) & (Dog, Gibbon) & (Dog, Lion) & (Dolphin, Elephant) & (Dolphin, Leopard) \\
             (Donkey, Gorilla) & (Donkey, Leopard) & (Donkey, Turtle) & (Donkey, Zebra) & (Duck, Fox) \\
             (Duck, Leopard) & (Duck, Monkey) & (Duck, Parrot) & (Duck, Potato) & (Duck, Wolf) \\
             (Eagle, Grapefruit) & (Eagle, Lizard) & (Eagle, Owl) & (Eagle, Parrot) & (Eagle, Tiger) \\
             (Earthworm, Snake) & (Elephant, Penguin) & (Elephant, Raccoon) & (Falcon, Fox) & (Falcon, Peacock)\\
             (Fish, Kiwi) & (Fish, Parrot) & (Fish, Seal) & (Fish, Shrimp) & (Fish, Tiger) \\
             (Fish, Wolf) & (Flamingo, Giraffe) & (Fox, Giraffe) & (Fox, Leopard) & (Frog, Hippopotamus)\\
             (Frog, Leopard) & (Frog, Tiger) & (Frog, Turtle) & (Frog, Whale) & (Frog, Wolf)\\
             (Garlic, Shark) & (Gibbon, Hippopotamus) & (Gibbon, Owl) & (Gibbon, Tiger) & (Giraffe, Peacock) \\
             (Giraffe, Squirrel) & (Gorilla, Lion) & (Gorilla, Lizard) & (Gorilla, Lobster) & (Hippopotamus, Horse) \\
             (Horse, Leopard) & (Horse, Shrimp) & (Kale, Octopus) & (Kale, Penguin) & (Kiwi, Owl) \\
             (Kiwi, Shark) & (Lemon, Octopus) & (Leopard, Lion) & (Leopard, Pig) & (Leopard, Rhinoceros) \\
             (Leopard, Squirrel) & (Leopard, Turtle) &(Leopard, Whale) & (Leopard, Wolf) & (Lettuce, Sheep) \\
             (Lettuce, Snail) & (Lion, Pineapple) & (Lion, Pumpkin) & (Lizard, Peacock) & (Lizard, Rhinoceros) \\
             (Lizard, Shark) & (Lizard, Turtle) & (Locust, Monkey) & (Locust, Peacock) & (Locust, Raccoon) \\
             (Mantis, Shark) & (Monkey, Owl) & (Monkey, Zebra) & (Moth, Octopus) & (Moth, Penguin) \\
             (Moth, Shrimp)& (Moth, Starfruit)& (Octopus, Peach) & (Octopus, Pineapple)&(Octopus, Potato)\\
             (Octopus, Raccoon) & (Octopus, Rambutan)&(Octopus, Squirrel)&(Octopus, Starfruit)&(Octopus, Watermelon)\\
             (Octopus, Zucchini)&(Ostrich, Owl) & (Owl, Parrot)&(Owl, Raccoon)&(Owl, Strawberry)\\
             (Peach, Penguin)&(Penguin, Pineapple)&(Penguin, Raccoon) &(Pig, Raccoon)&(Pig, Sheep)\\
             (Pig, Squirrel) & (Raccoon, Sheep)&(Raccoon, Spider)&(Raccoon, Walrus) &(Seal, Spider)\\
             (Shark, Wolf)&(Shrimp, Toad)&(Snail, Tiger)&(Snail, Watermelon)&(Snail, Wolf) \\
             (Snail, Zebra)&(Spinach, Wolf)&(Starfruit, Toad)&(Strawberry, Wolf)&(Toad, Turtle)\\
             \bottomrule[1pt]
        \end{tabular}
        }
    \label{tab:app_dataset}
\end{table*}

\end{CJK}
\end{document}